\pdfoutput=1

\documentclass[11pt]{article}

\usepackage{acl}

\usepackage{times}
\usepackage{latexsym}
\usepackage{hyperref}

\usepackage[T1]{fontenc}

\usepackage[utf8]{inputenc}

\usepackage{microtype}

\usepackage{inconsolata}
\usepackage{graphicx}
\usepackage{multirow}
\usepackage{booktabs}
\usepackage{tabularx}
\usepackage{amsmath}
\usepackage{nicematrix}

\usepackage{todonotes}
\usepackage[shortlabels,inline]{enumitem}
\setlist[itemize]{nosep,leftmargin=*}
\usepackage{cleveref}
\crefname{section}{s}{Sections}
\crefname{table}{Table}{Tables}
\crefname{figure}{Fig.}{Figs.}
\crefname{algorithm}{Alg.}{}
\crefname{ALC@unique}{Line}{Lines}
\crefname{equation}{Eq.}{Eqns.}
\crefname{appendix}{Appendix}{}
\crefformat{section}{\S#2#1#3}
\usepackage[normalem]{ulem}
\usepackage{graphicx}

\newcommand{\downnumNR}[1]{\textcolor{blue}{$\downarrow$#1}}
\definecolor{darkred}{RGB}{180,0,0}
\newcommand{\downnumGIST}[1]{\textcolor{darkred}{$\downarrow$#1}}
\definecolor{darkgreen}{RGB}{0,150,0}
\newcommand{\upnum}[1]{\textcolor{darkgreen}{$\uparrow$\footnotesize{#1}}}

\title{Beyond Math: Stories as a Testbed for Memorization-Constrained Reasoning in LLMs}

\author{Yuxuan Jiang , Francis Ferraro \\
Department of Computer Science and Electrical Engineering\\
  University of Maryland, Baltimore County \\
  \texttt{yuxuanj1@umbc.edu} }

\begin{document}
\maketitle
\begin{abstract}

Memorization has been shown to greatly inflate Large Language Models’ (LLMs) performance on domains such as math and logic, 
where success should primarily rely on applying generalizable reasoning rules. 
In many real-world applications, however, memorization is not meant to be eliminated but \textit{selectively constrained}—for example, in story understanding, where background knowledge must be integrated with narrative context. 
Drawing on the cognitive science distinction between “verbatim” (exact recall) and “gist” (semantic abstraction) memorization, 
we propose a two-tier framework for analyzing how LLMs reason under different degrees of memory access. 
The \textbf{Inductive (prompt-guided) Setting} softly steers models to reason through selective, context-relevant recall, 
while the \textbf{Restrictive Setting} imposes stronger constraints by limiting verbatim memory access. 
Evaluating GPT-4o, LLaMA3.3-70B, and DeepSeek V3 on six character-centric story understanding benchmarks, 
we find up to a 45.2\% accuracy drop under the Restrictive Setting, revealing strong dependence on surface recall. 
By contrast, the Inductive Setting maintains performance, indicating that prompting can align LLMs toward memorization-constrained reasoning.

\end{abstract}

\section{Introduction}

\begin{figure*}[h]
    \centering
    \includegraphics[width=0.95\linewidth]{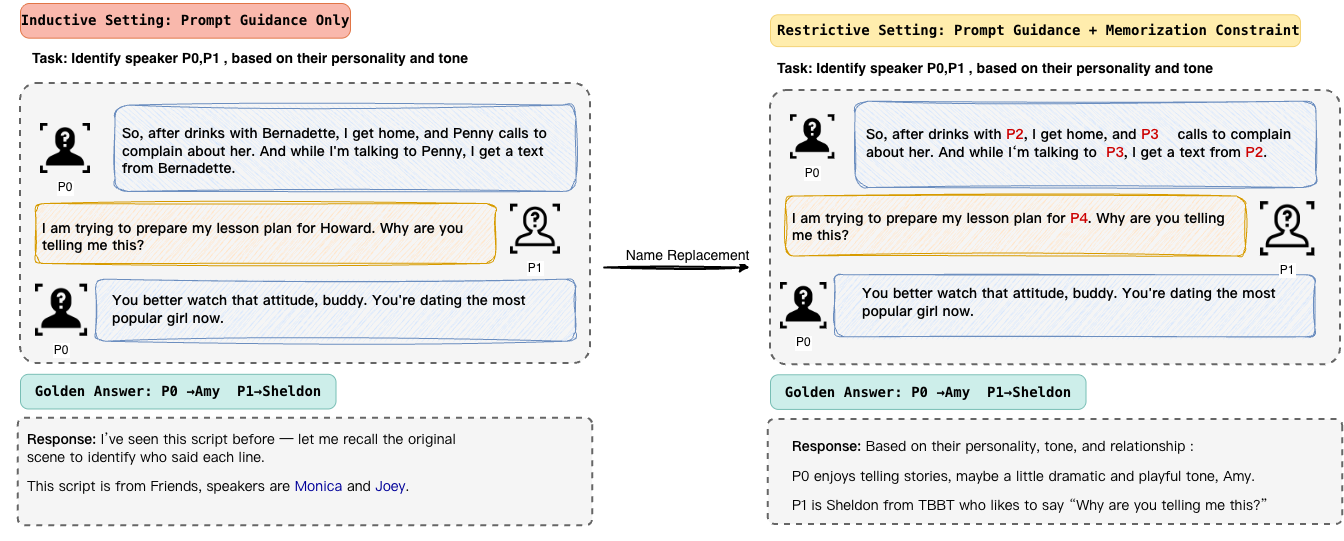}
    \caption{
Illustration of our graded intervention framework. 
The \textbf{Inductive Setting} encourages selective, context-based reasoning, 
while the \textbf{Restrictive Setting} limits verbatim recall by replacing character names.}

\label{GV}
\end{figure*}

Large Language Models (LLMs) have achieved remarkable success on a wide range of reasoning benchmarks. 
However, the manner in which these successes are achieved is often less satisfying. %
This can be especially salient in character-centric story understanding tasks, one of which we show in Figure~\ref{GV}: there, we see how a model can correctly solve the task of identifying a character in a dialogue. However, it does so by simply recalling the original script verbatim---a form of \textit{memorization} that deviates from the intended reasoning behavior. 
As we see, once the character names are replaced, this shortcut collapses, forcing the model to rely on contextual cues such as a ``dramatic and playful'' tone to recover the correct speaker through genuine reasoning. This transition under memorization constraints raises a deeper question: 
\textit{Can LLMs reason in the way humans expect---integrating selective background knowledge with narrative context rather than relying solely on surface recall?}%

Character-centric story understanding provides an ideal testbed for probing this issue. 
Such tasks demand combining background knowledge with story context to infer characters’ motivations, emotions, and relationships~\cite{yu_personality_2023,vallurupalli2024saga,shen_roleeval_2024,brahman_let_2021}. Because many benchmarks originate from well-known media sources, LLMs may have encountered them verbatim during pretraining~\cite{chang_speak_2023}. When characters’ names or stylistic cues are altered, memorization shortcuts collapse, exposing the model’s true reasoning capacity.

Prior work has examined the tension between memorization and reasoning mainly in domains such as math and logic, where models are expected to apply general rules and treat near-duplicates as new problems~\cite{lesci2024causal,jin2024disentangling,carlini_quantifying_2023,prabhakar2024deciphering,chen2024multi}. In contrast, many real-world applications require what we call \textit{memorization-constrained reasoning}, where memorization is not entirely undesirable but must be used \textit{selectively}---retrieving bounded, context-relevant knowledge and integrating it with local evidence. The challenge, therefore, is not to eliminate memory, but to understand how reasoning operates when memorization is constrained rather than removed.

To examine this systematically, we propose a \textbf{graded intervention framework} inspired by the cognitive distinction between  \textit{verbatim} (exact recall) and \textit{gist} (semantic abstraction) memory.  The \textbf{Inductive Setting} preserves the original text but introduces \textit{gist-based prompting} to encourage selective, context-driven recall---probing whether reasoning can be \textit{elicited} through guided induction.  The \textbf{Restrictive Setting} imposes a stronger constraint: all character names are replaced with culturally distinct alternatives, suppressing direct memory access to test whether reasoning can be \textit{sustained} when memorization is blocked. 
Together, these two settings form a continuum of interventions that distinguish between  \textit{behavioral steerability} (whether models can be guided to reason as intended) and \textit{intrinsic robustness} (whether reasoning endures under strict memory constraints).

Applying this framework to six character understanding tasks with three representative LLMs (GPT-4o, LLaMA3-70B, and DeepSeek V3), we find clear support for our hypothesis. 
Performance drops by as much as 45.8\% in the \textbf{Restrictive Setting}, showing that current benchmarks substantially overestimate reasoning by allowing verbatim memorization to dominate. 
By contrast, accuracy in the \textbf{Inductive Setting} remains largely stable, indicating that prompting can guide models toward reasoning that appropriately integrates memorization rather than relying on surface recall. 
Together, these findings reveal not only a critical gap in how current evaluations interpret “reasoning,” but also a concrete way to probe and improve it.

In summary, our key contributions are:

\begin{enumerate}
    \item \textbf{Expanding from Reasoning to Understanding Tasks:}  
    We extend the study of memorization and reasoning beyond math and logic to \emph{character-centric story understanding}, a domain that requires controlled use of memorization for reasoning.

    \item \textbf{An Inductive–Restrictive Framework:}  
    We propose a two-tier design: the \textbf{Inductive Setting} guides models toward memorization-constrained reasoning, while the \textbf{Restrictive Setting} limits verbatim recall to test intrinsic reasoning ability.

    \item \textbf{Diagnosing and Steering LLM Reasoning:}  
    Our results reveal that restricting verbatim cues causes large performance drops, while inductive prompting maintains accuracy—showing how prompting can steer models toward the intended, constrained form of reasoning.
\end{enumerate}


\section{Related Work}
\label{related}

\subsection{Memorization and Reasoning in LLMs}

LLMs’ impressive reasoning capabilities have attracted significant attention~\cite{tong2023eliminating,yang2023harnessing,xiong2024large,tan2024large,jiang2025drp,jiang2026scribe,xu2025learning,zhang2025find}, yet their behavior is also strongly shaped by memorization~\cite{zhou_dont_2023,chen2024multi,biderman2024emergent}. 
Prior studies show that memorizing training data can inflate benchmark scores~\cite{khandelwal_generalization_2020,blevins_language_2022} and that conflicting memories can even degrade performance~\cite{su_conflictbank_2024,wang_resolving_2024,yang2025quantifying}. 

Cognitive science offers a useful lens for this distinction, separating \textit{verbatim memorization}—precise recall of surface details—from \textit{gist memorization}—semantic abstraction of meaning~\cite{reyna1998fuzzy,brainerd2002fuzzy}. 
To avoid confusion with architectural notions of memory in NLP~\cite{alizadeh2024llm,kwon2023efficient}, we use these terms to describe distinct \textit{recall styles} rather than neural mechanisms. 
This distinction underpins our analysis of how LLMs balance shallow recall with deeper reasoning.

Recent NLP research has sought to disentangle and quantify memorization~\cite{lesci2024causal,jin2024disentangling,carlini_quantifying_2023,prabhakar2024deciphering}, examining phenomena such as multifaceted memorization~\cite{prashanth2024recite,li2024romememorizationinsightstext} and logit-based tracing~\cite{chen2024multi}. 
While mitigation methods like out-of-distribution filtering~\cite{biderman2024emergent} help in structured reasoning tasks, they are less applicable to character-centric domains grounded in known fictional texts. 
In contrast, we propose a prompting-based framework that \textit{tests whether LLMs can regulate their use of memorization under explicit constraints}, distinguishing verbatim recall from reasoning that integrates bounded, context-relevant memory.

\subsection{Character centric Story Understanding and Its Evaluation with LLMs}

Character understanding has long been a focus in psychological, literary, and educational research~\cite{bower1990mental, mckee2010story, paris2003assessing}. It spans tasks such as character recognition, coreference resolution, summarization, role detection, goal modeling, personality analysis, and question answering~\cite{ bamman_annotated_2020, chen_summscreen_2022, frermann_whodunnit_2018, vallurupalli2024saga, yu2023personality}. 

Recent work demonstrates that LLMs achieve strong results on character-centric benchmarks~\cite{li2023multi, stammbach_heroes_2022, goyal_news_2023}. However, many of these datasets are derived from publicly available or widely consumed fictional sources, such as movie scripts, TV shows, and novels~\cite{sang2022survey, tapaswi_movieqa_2016, caselli_event_2017}, which makes them vulnerable to pretraining overlap and surface-level memorization~\cite{chang_speak_2023}. Although prior research has studied LLM memorization broadly~\cite{carlini_quantifying_2023, xie2024memorization,yang2023hallu}, little is known about how it specifically affects character understanding tasks. Our work fills this gap by examining memorization's role in these tasks and proposing  interventions to distinguish between recall and reasoning.

\section{Data and Experiment Setup}
\label{data}

To comprehensively evaluate character understanding in LLMs, we compile two datasets: \textbf{CharBench} and \textbf{CharScript}. Inspired by NLP benchmark collections such as GLUE~\cite{wang2018glue}, our \textbf{CharBench} integrates six publicly available datasets spanning character-centric tasks like coreference resolution, personality inference, and dialogue-based QA (details in Section~\ref{6tasks}).

The second dataset, \textbf{CharScript}, is newly constructed for this study by systematically sampling textual segments from original fictional sources underlying CharBench, explicitly probing models' verbatim memorization (details in Section~\ref{fic memorization}).

\subsection{CharBench: A Benchmark Collection for Character Understanding}
\label{6datas}

CharBench aggregates six widely recognized NLP benchmarks covering diverse narrative mediums—TV shows, movies, and novels. To ensure comparability, we retain each dataset’s original splits, preprocessing, and evaluation metrics. Detailed task descriptions and implementations are provided in Appendix~\ref{benchdata}.

\paragraph{Benchmark Sources}  
Following established NLP benchmark practices~\cite{wang2018glue}, CharBench includes:

\begin{itemize}[leftmargin=1.2em]
    \item \textbf{MovieCoref}~\cite{chen2016character}: Coreference resolution in movie scripts.
    \item \textbf{TVSHOWGUESS}~\cite{sang2022tvshowguess}: Speaker identification in TV dialogues.
    \item \textbf{PERSONET}~\cite{yu2023personality}: Personality inference from literary passages.
    \item \textbf{CSI Corpus}~\cite{frermann_whodunnit_2018}: Role detection from crime shows.
    \item \textbf{FriendsQA}~\cite{yang_friendsqa_2019}: Multi-turn dialogue-based QA.
    \item \textbf{SummScreen}~\cite{chen_summscreen_2022}: TV episode summarization.
\end{itemize}

\paragraph{Evaluation Metrics} 
We strictly follow original dataset metrics (exact match, ROUGE, F1 score), as detailed in Table~\ref{benchmark_datasets}.

\begin{table}[t]\small
\centering
\setlength{\tabcolsep}{1.2mm}
\resizebox{0.48\textwidth}{!}{
\begin{tabular}{l l l l}
\toprule
\textbf{Task Name} & \textbf{Task Type} & \textbf{Data Source} & \textbf{Metric} \\
\midrule
TVShow Guess   & Character Guessing         & TV Shows       & Exact Match \\
FriendsQA         & Question Answering   & TV Shows       & Exact Match \\
Screenplay        & Character Coreference  & Movies         & F1 Score \\
SummScreen        & Text Summarization               & TV Shows           & ROUGE \\
CSI               & Role Detection                  & TV Shows       &   F1 Score \\
PERSONET          & Personality Understanding             & Novels         & Exact Match\\
\bottomrule
\end{tabular}
}
\caption{Overview of benchmark tasks, including their type, data sources, and evaluation metrics.}
\label{benchmark_datasets}
\end{table}

\subsection{CharScript: Fictional Works Memorization Benchmark Dataset}
\label{fic memorization}

Motivated by recent NLP research on memorization in language models~\cite{chang2023speak}, we constructed the CharScript dataset to evaluate whether LLMs can recognize the source of fictional text segments. Following a similar setup to~\cite{chang2023speak}, we probe whether models can correctly attribute short passages to their original narrative sources, as an indicator of verbatim memorization. We present our experimental design and findings in Section~\ref{memorization eval}.

We systematically sampled and extracted \textbf{200 textual segments}, averaging approximately \textbf{312 words each}, from \textbf{40 diverse fictional sources} (11 TV series, 9 movies, and 20 novels). These sources were carefully selected to maximize narrative diversity, textual complexity, and representativeness of popular fictional works included in CharBench. Each source contributes exactly \textbf{5 segments}, structured uniformly in the format \textit{[Title]; Scene context}, combining both original dialogue and narrative descriptions.

\subsection{Models}
We evaluate three frontier LLMs—open-source LLaMA 3.3-70B~\cite{meta2024llama3.3} and DeepSeek V3~\cite{liu2024deepseek}, as well as the closed-source GPT-4o~\cite{openai2023chatgpt}—via API-based inference. To address stochastic variability, each model was run three times per setting, and average performance is reported. The total cost of our experiments was approximately \$100.


\section{Inductive Setting: Guiding Memorization-Constrained Reasoning}
\label{soft setting}

In this setting, we aim to examine whether LLMs can follow explicit prompt instructions 
to regulate their use of memorization and reasoning as intended. 
Specifically, we test whether prompting alone—without altering the input text—can 
induce models to move away from verbatim recall and instead engage in reasoning 
that relies on relevant, context-bound memory and semantic abstraction.

\subsection{Character Guessing Task Format}

We select the Character Guessing Task specifically because it closely mirrors the name cloze tests~\cite{chang2023speak} previously established as effective probes for verbatim memorization in LLMs. Name cloze tests measure memorization by removing character names from familiar contexts and testing if models can accurately recall them. Similarly, in our task, speaker labels are removed from dialogue excerpts, requiring the model to infer speakers solely from textual context. If models rely heavily on verbatim memorization, accuracy should significantly degrade when surface cues—such as explicit character names—are disrupted. Conversely, if models genuinely reason about character interactions and relationships, performance should remain relatively stable despite these perturbations.

Following the original TVShow Guess setup~\cite{sang_tvshowguess_2022}, the model is tasked with predicting the speaker of each utterance in a dialogue segment where all speaker labels are removed.

\subsection{Prompting Strategies}
\label{promptss}
We design two distinct prompting conditions to examine how LLMs rely on different memorization mechanisms when identifying characters.  
Verbatim prompting encourages mechanical memorization, where the model recalls exact substrings from its training data and reconstructs missing parts without reasoning. In contrast, gist prompting focuses on extracting key information and performing memorization-based matching. For character understanding tasks, the most relevant gist cues typically include character relationships, significant events, and personality traits.  

All prompts share the same base instruction:  \textit{"The following is a dialogue. Please identify who they are. …"}

The two variations are as follows:

\textbf{Verbatim Prompt:}  
\textit{"...Do this by direct memorization retrieval—do not infer based on traits, interactions, or context. Recall names exactly from past dialogue knowledge without explanations or reasoning."}  

\textbf{Gist Prompt:}  
\textit{"...Do this by analyzing relationships, key events, and personality traits to match them with known characters. Do not directly retrieve from memorization who said the sentences."}

\paragraph{Prompt Robustness}
To positively establish that our findings are stable across wording choices, we designed two paraphrastic variants for each condition (\emph{verbatim} and \emph{gist}), keeping the instructional structure fixed while slightly altering phrasing (Appendix~\ref{app:prompt-robust}). For each model, we ran three trials per variant and report mean accuracy. Across all models, the qualitative ordering observed in the main results is consistently reproduced under these nearby formulations. Numerical results appear in Appendix Table~\ref{tab:minimal_paraphrase}.

\subsection{Result and Analysis}
\label{softresult}

Our key observations are as follows:\\
\begin{itemize}
    \item \textbf{Baseline Performance:} The baseline condition reflects the overall performance of the models, with GPT-4o demonstrating a clear lead to others, while Llama 3.3-70B and DeepSeek V3 exhibit quite similar performance.
    
    \item \textbf{Memorization Reliance:} All three models show that the \textit{Verbatim} condition performs nearly identically to the baseline, supporting our hypothesis that models primarily rely on memorization when answering the \textit{Character Guessing} task.
    
    \item \textbf{Effect of Gist Prompting:} The \textit{Gist} condition results in a non-trivial performance drop, approximately 10\%, suggesting that our prompt effectively reduces the model's dependence on memorization when generating responses.
\end{itemize}

\begin{figure}[!t]
    \centering
    \includegraphics[width=7.5cm]{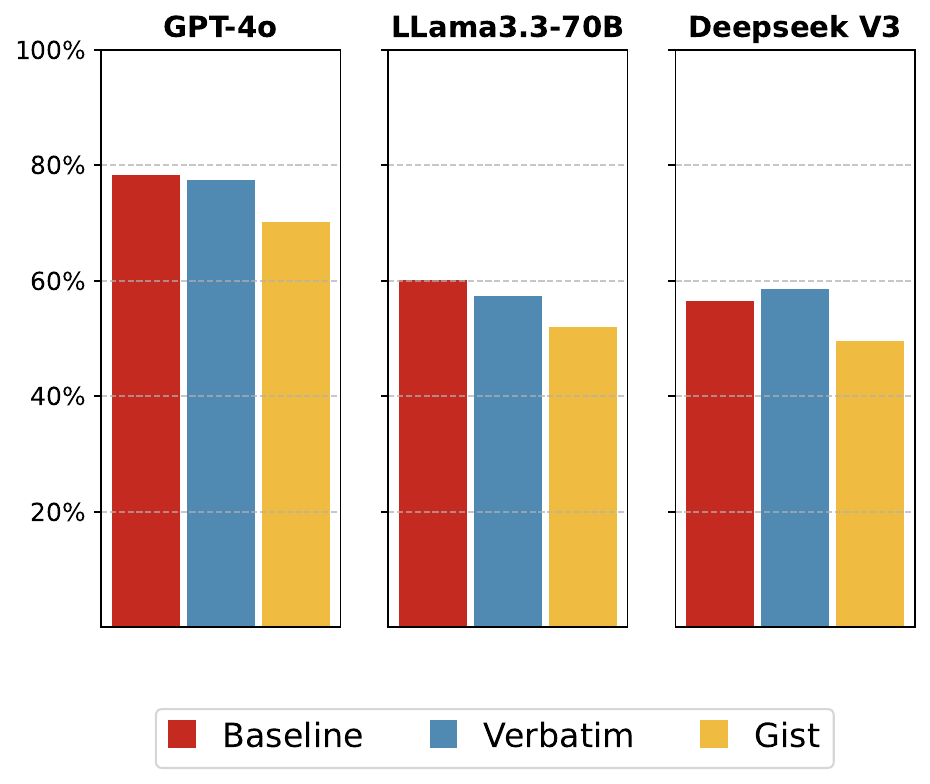}
    \caption{The bar chart compares the performance of three LLMs under three prompting settings. Across all three models, the performance of the Verbatim and Baseline conditions is highly similar, confirming our hypothesis that the models rely on memorization when answering questions. The Gist condition, however, shows a non-trivial decline in performance, indicating that the models struggle when they cannot fully rely on memorization to generate responses.} 
    \label{soft}
\end{figure}

Our results indicate that gist-based prompting can indeed shift model responses toward reasoning, as evidenced by the performance drop under the \textit{Gist} condition (see Figure~\ref{soft}). However, the moderate performance drop (approximately 10\%) suggests that even gist prompting may not completely eliminate the models' reliance on verbatim memorization. Previous research similarly indicates that prompting without explicit reasoning steps may fail to fully engage true reasoning processes~\cite{yu2025llmsreallythinkstepbystep}. Therefore, to conclusively test our hypothesis—that models' strong benchmark performance relies significantly on verbatim memorization—we introduce a more direct and explicit intervention aimed at disrupting memorized surface-level cues.

\section{Restrictive Setting: Testing Reasoning under Memorization Constraints}
\label{task2}

We further test whether models can reason effectively when direct memorization shortcuts are blocked. 
In this setting, all character names are replaced with culturally distinct alternatives, 
removing verbatim recall while preserving narrative meaning. 
This allows us to evaluate whether LLMs can perform reasoning 
when surface-level memorization cues are no longer accessible.

\subsection{Name Replacement Strategies and Their Impact on Memorization}

\paragraph{Name Replacement Strategies}\label{nrs} Previous studies~\cite{yu2022few} have primarily employed Same-Cultural Name Replacement, a naive and intuitive approach that substitutes character names with culturally consistent alternatives. To systematically analyze the impact of name replacement on model performance, we introduce three distinct levels of name modifications:

\begin{itemize}
    \item \textbf{Name Masking:} Replacing character names with placeholders (e.g., \textit{Sheldon, Amy} → P0, P1). 
    \item \textbf{Cross-Cultural Name Replacement:} Substituting names with those from a different cultural background (e.g.English → Chinese) while maintaining gender consistency (e.g., \textit{Sheldon, Amy} → \textit{Jie Zhang, Xiaoling Wang})~\cite{sun2024fostering}. This tests whether models rely on specific name distributions for character inference.
    \item \textbf{Same-Cultural Name Replacement:} Replacing names with other culturally consistent alternatives (e.g., \textit{Sheldon, Amy} → \textit{Andrew, Sally}). This retains cultural familiarity while disrupting direct memorization by name replacements.
\end{itemize}

\paragraph{Human Validation of Reasoning Preservation}
We conducted two complementary human validations (Appendix~\ref{human}). \emph{Exp.~1 (Objective accuracy)} examines whether our name replacement \emph{itself} reduces accuracy; with the accuracy remains no obvious change (max $|\Delta|=2.2$ percentage points and no item was judged “unsolvable.”), implying that any model-side drop under replacement is attributable to reliance on verbatim memorization. \emph{Exp.~2 (Gold/semantic integrity)} tests whether replacement changes or ambiguates the gold answer; across 300 item–judgments, none were flagged. Together, these results show that name replacement preserves the semantic structure of the items and keeps gold labels intact.

\paragraph{Memorization Evaluation Task Format} 
\label{memorization eval}

To directly quantify how much verbatim memorization models rely on, we design a \textbf{Source Identification Task}. This task closely resembles established memorization evaluation methods, such as source attribution or textual cloze tests~\cite{chang2023speak}. Specifically, given a dialogue excerpt \( D = \{U_1, U_2, ..., U_n\} \) without source annotations, the model must identify the correct origin of the text, predicting \( f_{\text{source}}(D) \rightarrow S \), where \( S \) is the specific fictional source.  

The rationale behind this design is straightforward: if a model has memorized verbatim text from pretraining, it should accurately identify the source based purely on surface-level textual recall. Conversely, if name replacement or prompting disrupts memorization effectively, we expect a significant decrease in source-identification accuracy. Thus, observing substantial performance degradation would strongly support the claim that models' prior accuracy was driven largely by verbatim memorization rather than reasoning. We use the \textbf{CharScript: Fictional Works Memorization Benchmark Dataset} described in Section~\ref{fic memorization} to rigorously evaluate this effect.

\begin{figure}[!t]
    \centering
    \includegraphics[width=7.5cm]{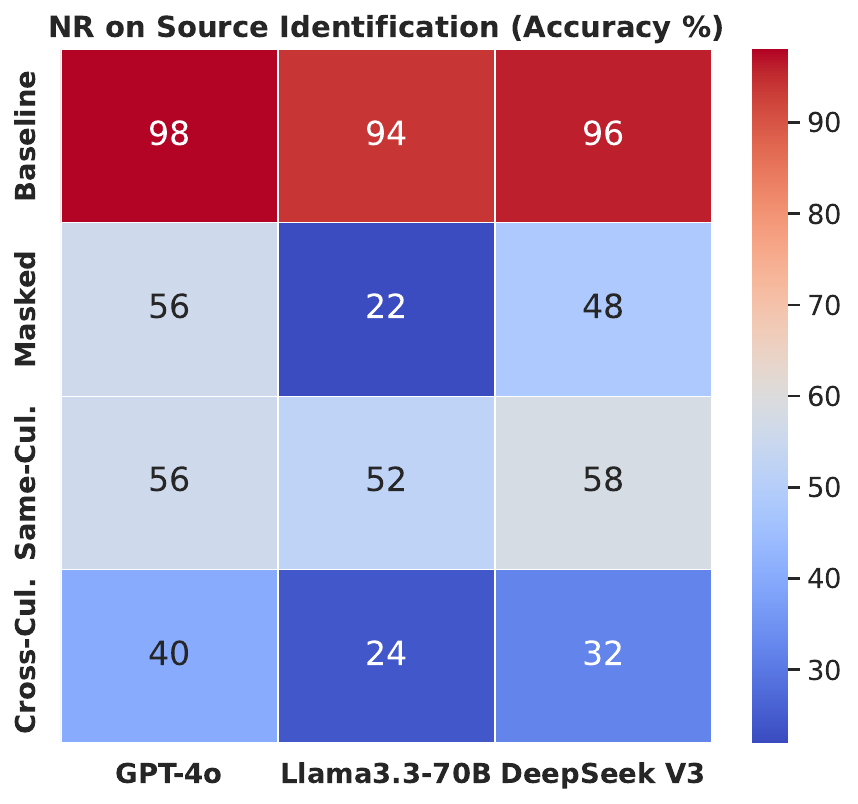}
    \caption{The heatmap presents accuracy percentages for source identification across three LLMs under different name replacement conditions. All three models demonstrate strong memorization of the script. The Cross-Culture Name Replacement method exhibits a significantly strong and the most pronounced memorization-weakening effect across all three models.} 
    \label{heat}
\end{figure}

\paragraph{Impact of Name Replacement on Memorization.}  
According to Figure~\ref{heat}, our results demonstrate that different name replacement strategies have a significant impact on the memorization ability of LLMs. Among them, Cross-Cultural Name Replacement consistently induces the strongest forgetting effect across all models, indicating that altering names to those from a different cultural background disrupts memorized associations more effectively than other replacement methods. This suggests that cultural distance may play a crucial role in weakening memorized identity-linking patterns in models.

\subsection{Memorization-Controlled Prompting under Name Replacement}
 
Different name replacement strategies have shown a significant impact on the memorization ability of LLMs. Building on this, we apply the name replacement strategies introduced in Section~\ref{nrs} to evaluate the \textit{Character Guessing} tasks, using the verbatim and gist prompting strategies we have discussed in Section~\ref{promptss}.  

Under the Name Replacement condition, we provide brief character descriptions generated by GPT-4, aiming to enable the model to complete the Character Guessing task through character matching. The descriptions used are provided in the Appendix~\ref{prompt}. To assess the actual effectiveness of these descriptions, we conduct an ablation study.

\begin{figure}[!t]
    \centering
    \includegraphics[width=8cm]{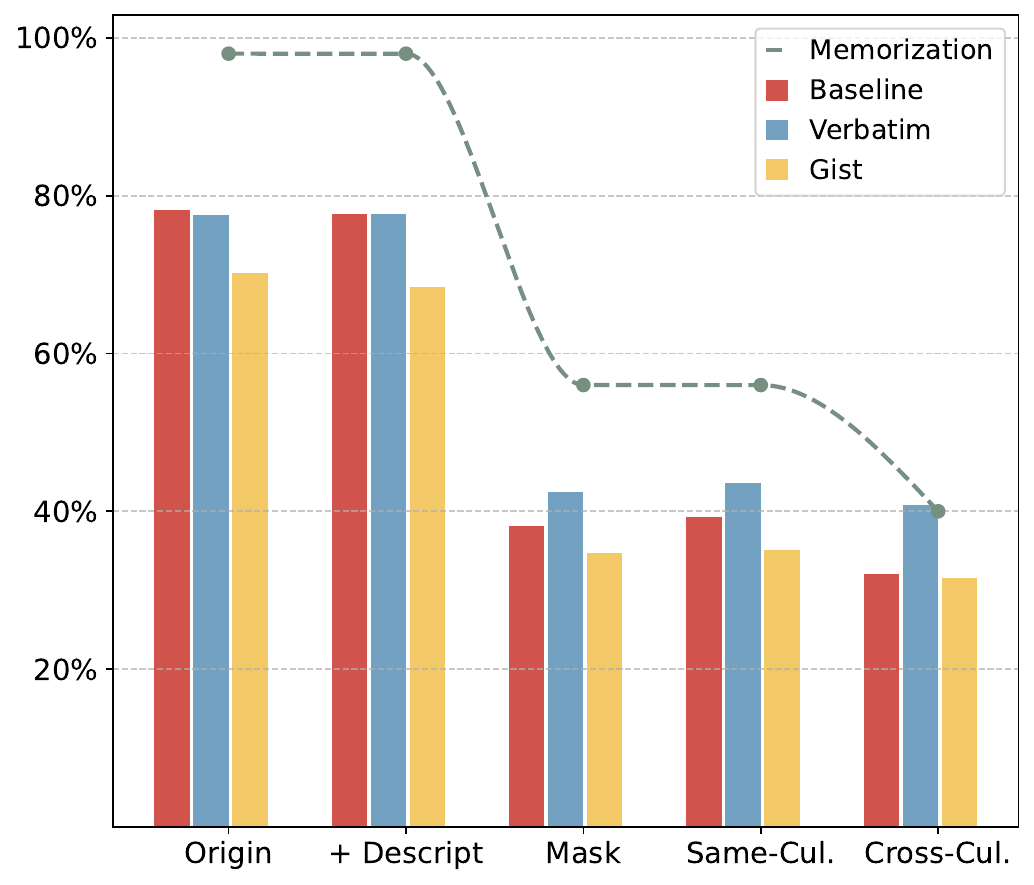}
    \caption{The line chart shows the decline in memorization across different settings, illustrating that as memorization weakens, overall model performance also decreases. The bar chart presents the effect of different prompting strategies: under Name Replacement conditions, Verbatim prompting consistently outperforms the baseline by reinforcing memorization, whereas Gist prompting further reduces reliance on memorization, leading to the lowest accuracy.}
 
    \label{nrfig}
\end{figure}

\paragraph{Results and Key Findings} 
We present the core results in Figure~\ref{nrfig} and summarize four key findings:

\begin{enumerate}
    \item \textbf{Performance Declines with Reduced Memorization.}  
    Accuracy drops as surface cues are removed, revealing the model's strong reliance on verbatim memorization.

    \item \textbf{Character Descriptions Have Minimal Impact.}  
    Adding character descriptions does not notably change performance, suggesting limited influence on model decisions.

    \item \textbf{Prompting Style Matters.}  
    Verbatim prompts boost performance by reinforcing memorization, while Gist prompts suppress it and lead to lower accuracy.

    \item \textbf{Name Replacement + Gist Prompting Is Effective.}  
    This combination best reduces memorization reliance and reveals the model’s true reasoning ability.
\end{enumerate}

\section{Evaluating Understanding under Memorization Constraints}
\label{task3}

We further evaluate whether our framework generalizes across diverse understanding tasks. 
Combining inductive prompting with restrictive name replacement, 
we test if LLMs can sustain reasoning performance when memorization cues are systematically constrained.

\subsection{Benchmark Tasks Description}
\label{6tasks}
We benchmark five types of character understanding tasks spanning coreference, personality inference, role detection, QA, and summarization. Each task evaluates the model's ability to extract or infer character-related information from dialogue.

\paragraph{Example Dialogue}

To illustrate these tasks, consider the following conversation:
\begin{quote}
\textbf{P0}: "You're still going to have to convince a jury that I killed two strangers for no reason."

\textit{Grissom doesn't look worried. He takes his gloves off and puts them on the table.}

\textbf{P1}: "You ever been to the theater, Peter? There's a play called Six Degrees of Separation."
\end{quote}

The correct speaker labels are \textbf{P0 → Peter Berglund, P1 → Grissom}. For all tasks except Character Guess, the placeholders P0 and P1 are replaced with their original names, Peter Berglund and Grissom. Table~\ref{task_examples} illustrates how this dialogue segment is utilized across different tasks.

\begin{table}[t]
\centering
\setlength{\tabcolsep}{5pt}
\renewcommand{\arraystretch}{1.3} 
\resizebox{0.48\textwidth}{!}{
\begin{tabular}{p{3.5cm} p{7.5cm}}
\toprule
\textbf{Task} & \textbf{Example Question and Expected Output} \\
\midrule

\textbf{Coreference Resolution} & "Which entity does 'his' refer to?" → \textbf{"Grissom"} \\  
\textbf{Personality Understanding} & "What trait best describes P1?" → \textbf{Analytical} \\  
\textbf{Role Detection} & "Identify all entities referring to the criminal." → \textbf{Peter Berglund} \\  
\textbf{Question Answering} & "What play does Grissom mention?" → \textbf{"Six Degrees of Separation"} \\  
\textbf{Summarization} & "Generate a summary." → \textbf{"Peter discusses convincing a jury, while Grissom references a play."} \\  
\bottomrule
\end{tabular}
}
\caption{Examples of different character understanding tasks applied to the same dialogue.}
\label{task_examples}
\end{table}

\subsection{Gist Prompts Across Character Understanding Tasks}

\paragraph{General Guidelines for Gist Prompting} \ 
\begin{itemize}
    \item \textbf{Encourage Structural Reasoning:} Guide models to infer character identity and relationships from discourse patterns rather than relying on explicit name associations.
    \item \textbf{Eliminate Memorization Bias:} Ensure prompts emphasize evidence-based inference rather than knowledge retrieval.
    \item \textbf{Preserve Narrative Coherence:} In summarization and question answering, focus on maintaining logical flow rather than phrase-matching.
\end{itemize}

Detailed prompts in Appendix~\ref{gistprompts}.

\begin{table*}[h]
    \centering
    \resizebox{\linewidth}{!}{%
    \begin{NiceTabular}{l|ccc|ccc|ccc}
        \toprule
        \multirow{2}{*}{\textbf{Tasks}} & \multicolumn{3}{c}{\textbf{GPT-4o}}  & \multicolumn{3}{c}{\textbf{LLaMA3.3-70B}} & \multicolumn{3}{c}{\textbf{DeepSeek V3}} \\
        \cline{2-10} & \textbf{Origin} & \textbf{NR} & \textbf{NR+GIST} & \textbf{Origin} & \textbf{NR} & \textbf{NR+GIST} & \textbf{Origin} & \textbf{NR} & \textbf{NR+GIST} \\
        \midrule
        Tvshow Guess & 78.2 & 32.0 \downnumNR{46.2} & 31.5 \downnumGIST{0.5} & 61.4 & 31.9 \downnumNR{29.5} & 27.8 \downnumGIST{4.1} & 60.8 & 29.7 \downnumNR{31.1} & 26.7 \downnumGIST{3.0} \\
        Coreference & 58.3 & 49.6 \downnumNR{8.7} & 47.5 \downnumGIST{2.1} & 56.7 & 48.3 \downnumNR{8.4} & 48.1 \downnumGIST{0.2} & 47.4 & 34.6 \downnumNR{12.8} & 32.3 \downnumGIST{2.3} \\
        FriendsQA & 44.3 & 38.6 \downnumNR{5.7} & 38.6 \upnum{0.0} & 45.7 & 42.9 \downnumNR{2.8} & 44.3 \upnum{1.4} & 42.8 & 39.4 \downnumNR{3.4} & 34.3 \downnumGIST{5.1} \\
        ScreenSum & 37.5 & 28.7 \downnumNR{8.8} & 23.1 \downnumGIST{5.6} & 40.1 & 25.9 \downnumNR{14.2} & 21.3 \downnumGIST{4.6} & 35.6 & 19.3 \downnumNR{16.3} & 17.1 \downnumGIST{2.2} \\
        CSI Role Extract & 50.1 & 32.9 \downnumNR{17.2} & 31.7 \downnumGIST{1.2} & 48.3 & 34.7 \downnumNR{13.6} & 32.7 \downnumGIST{2.0} & 45.1 & 29.8 \downnumNR{15.3} & 23.7 \downnumGIST{6.1} \\
        PERSONET & 56.0 & 52.0 \downnumNR{4.0} & 54.0 \upnum{2.0} & 62.0 & 46.0 \downnumNR{16.0} & 44.0 \downnumGIST{2.0} & 44.0 & 40.0 \downnumNR{4.0} & 38.0 \downnumGIST{2.0} \\
        \bottomrule
    \end{NiceTabular}
}
    \caption{Ablation study examining the impact of name replacement and the GIST method on model performance across six NLP tasks.The subscripted values in the \textbf{NR} column indicate the performance drop relative to the original script, while the subscripted values in the \textbf{NR+GIST} column show the performance change compared to the Name Replace condition. }
    \label{performance_comparison}
\end{table*}
\subsection{Results and Key Findings}

Our evaluation provides new insights into how LLMs adapt their reasoning under memorization constraints, summarized in three key findings:

\begin{enumerate}
    \item \textbf{Models Default to Verbatim Recall but Can Adapt under Constraints.}  
    As shown in Table~\ref{performance_comparison}, restricting surface cues through name replacement consistently reduces accuracy, revealing that models initially rely on verbatim recall.  
    However, when guided by inductive prompting, performance partially recovers, suggesting that LLMs can adjust toward reasoning supported by bounded, context-relevant memorization rather than full recall.

    \item \textbf{Task Type Influences Memorization Dependence.}  
    Action-oriented tasks (e.g., Character Guessing, CSI Role Detection) depend more on direct cue–label associations, making them sensitive to memorization constraints.  
    In contrast, motivation-oriented tasks (e.g., Personality Understanding) encourage inference about internal states and intentions, naturally aligning with reasoning that integrates but does not overuse memory.

    \item \textbf{Excessive Memorization Can Impair Reasoning.}  
    In some benchmarks, models perform better once memorization is constrained, as removing misleading associations alleviates knowledge conflicts.  
    This indicates that controlled memory use—rather than maximal recall—supports more faithful character understanding.
\end{enumerate}

\begin{figure}[!t]
    \centering
    \includegraphics[width=8cm]{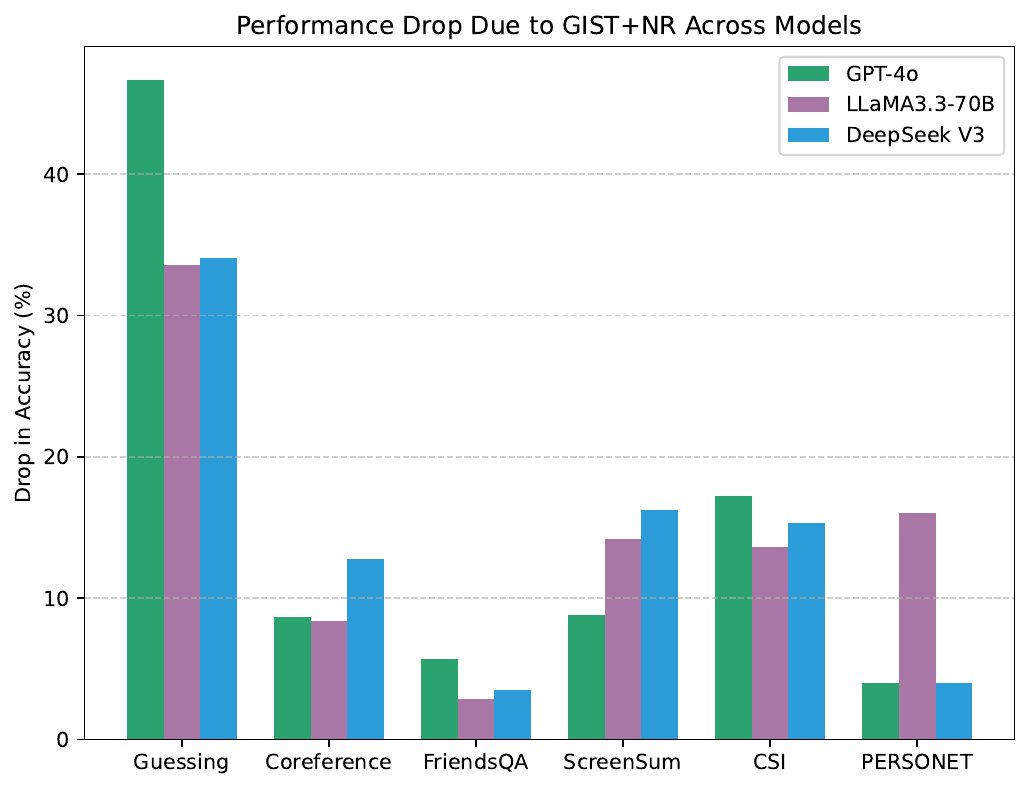}
    \caption{Performance degradation across LLMs due to GIST+NR. 
Action prediction tasks (e.g., \textit{Guessing}, \textit{CSI}) show greater drops than motivation prediction tasks (e.g., \textit{PERSONET}, \textit{FriendsQA}), reflecting their differential reliance on memorization.}
    
    \label{drops}
\end{figure}

These findings are consistent across multiple models (Figure~\ref{drops}), suggesting generalizability and broader implications of memorization biases in current benchmarks.

\section{Conclusion}

We examined how LLMs use memorization in memorization-constraint tasks, framing it not as a flaw but as a process that should be constrained and integrated with reasoning. 
Our two-tier framework—the \textbf{inductive} and \textbf{restrictive} settings—tests whether models can regulate memorization when guided by prompts or deprived of surface cues. 
Results across six character-centric benchmarks show that while models default to verbatim recall, they can be steered toward reasoning under bounded memory. 
This highlights the need for evaluations that assess reasoning under realistic memorization constraints.

\section{Acknowledgments}
We wish to thank the anonymous reviewers for their helpful comments, feedback, and suggestions. %
Some experiments were conducted on the UMBC HPCF, supported by the National Science Foundation under Grant No. CNS-1920079. %
This material is also based on research that is in part supported by DARPA for the SciFy program under agreement number HR00112520301. The U.S. Government is authorized to reproduce and distribute reprints for Governmental purposes notwithstanding any copyright notation thereon. The views and conclusions contained herein are those of the authors and should not be interpreted as necessarily representing the official policies or endorsements, either express or implied, of DARPA or the U.S. Government.



\section{Limitations}
The specific training data of GPT-4o is not publicly disclosed by OpenAI, making it fundamentally unknowable. Our work employs probabilistic inference to estimate the model's familiarity with a set of fictional works. However, whether these works are explicitly included in the training data remains unanswerable. Nevertheless, even mere familiarity can suffice to contaminate test data, posing challenges to fair evaluation.

Our study can be extended by future research in a number of ways. First, though we ground and motivate our work through ``verbatim'' and ``gist'' memorization, we note that any apparent ``reasoning'' of AI systems may differ fundamentally from those of humans, and we did not pursue human studies to examine this in detail. Further investigation is needed to explore this topic.

Second, the relationship between memorization and reasoning remains a subject of debate in cognitive science. Future studies may uncover insights into this relationship from entirely new perspectives.

Thirdly, While our interventions reveal strong trends, we acknowledge that results may be sensitive to the surface form of prompts. Future work could complement our findings with uncertainty-based analyses such as log probability comparisons.

Finally, our study was centered on select, recent (but not contemporaneous) majority Western popular culture references. Given their popularity, this selection likely increases the chances that LLMs have been pretrained on more instances or language discussing those references (vs. contemporaneous references, or references from non-Western cultures, where overall occurrences in training data may be lower); it is possible that this exaggerated some of the effects we observed.

\section{Ethics}
Our study probes the degree to which GPT-4o have memorized fictional works, and its impact on character understanding tasks. This work uses the OpenAI API on experiments above, and at no point do we access, or attempt to access, the true training data behind these models, or any underlying components of the systems.
\textbf{Risks}
The several datasets and fictional work scripts in our experiment are sourced from publicly available sources. However, we cannot guarantee that they are free from socially harmful or toxic language. We use ChatGPT~\footnote{\url{https://chatgpt.com/}} to correct grammatical errors in this paper.
\paragraph{Annotator Characteristics.} 
The three participants were adult graduate students located in the United States. 
Two had no prior exposure to the target TV series, while one was familiar with it. All volunteers, no payment. 
No additional demographic attributes were collected. Instructions given can be found in Appendix~\ref{human}.

\paragraph{Consent and Data Use.} 
All participants were informed that the study involved answering comprehension questions 
for research purposes. They were told that their responses would be anonymized, used only 
for validation analysis, and not shared beyond the scope of this paper. 
Written consent was obtained before participation.

\bibliography{anthology,custom}
\bibliographystyle{acl_natbib}

\appendix

\section{Example of Script-based Character Understanding Task}
\label{Example of Script-based Character Understanding Task}

In \cref{example: character guessing} we show a full example of the character understanding task. In this example, we have automatically replaced the individual character names with placeholders like P0 and P1.

\begin{table*}[h]
\centering
\begin{tabular}{lp{4.5in}p{2.9in}}
\hline
Input & P0: Hey, sorry about that P1: No, we're sorry. We never should have been comparing relationships in the first place. P2: Why? We won. You know, I say, next, we take on Koothrappali and his dog. Really give ourselves a challenge. P3: I just want to say one more thing about this. Just because Penny and I are very different people does not mean that we're a bad couple. P2: The answer is one simple test away. Hmm? You know, it's like when I thought there was a possum in my closet. Did I sit around wondering? No, I sent Leonard in with a pointy stick and a bag. P3: I killed his Chewbacca slippers. P0: Let's just take the test. P3: No, no, no, I don't want to. P0: Oh, well, 'cause you know we're gonna do bad. P3: Because it doesn't matter. I don't care if we're a ten or a two. P2: Or a one. A one is possible. P3: Marriage is scary. You're scared, I'm scared. But it doesn't make me not want to do it. It, it just makes me want to hold your hand and do it with you. P0: Leonard. P1: It makes me so happy if you said things like that. P2: We got an eight-point-two. Trust me, you're happy. \\ \hline
Label & P0: Penny, P1: Amy, P2: Sheldon, P3: Leonard                                                                         \\ \hline
\end{tabular}
\caption{An example from character guessing task.}
\label{example: character guessing}
\end{table*}

\begin{table}[t]
\centering
\begin{tabular}{lp{1.8in}p{2.3in}}
\hline
Input & Tell me the source for following script:Penny: Hey, sorry about that Amy: No, we're sorry. We never should have been comparing relationships in the first place. Sheldon: Why? We won. You know, I say, next, we take on Koothrappali and his dog. Really give ourselves a challenge. Leonard: I just want to say one more thing about this. Just because Penny and I are very different people does not mean that we're a bad couple.  \\ \hline
Source & The Big Bang Theory                                                     \\ \hline
\end{tabular}
\caption{An example for source prediction task.}
\label{namepre}
\end{table}

\section{Extended Dataset Analysis}
For the six representative character understanding tasks studied in our work, we selected one dataset for each category. However, we also recognize that there are many different benchmark datasets for each type of task. When selecting representative datasets for each category, we focused on choosing those with more diverse sources. Additionally, we supplemented the selection with eight datasets of the same type but derived from different sources in Table~\ref{extended_tasks}.

\begin{table}[t]
\centering
\setlength{\tabcolsep}{0.9mm}
\resizebox{0.98\columnwidth}{!}{
\begin{tabular}{lccc}
\toprule
Task Name            & Task Type                    & Source           \\
\midrule
Character Mining     & Character Identification     & TV Shows          \\
LISCU                & Character Identification     & Book Reviews      \\
MovieQA              & Visual Question Answering    & Movies            \\
Liter Coreference    & Character Coreference        & Novels            \\
HVV Detect           & Role Detection              & Movies, News      \\
SYMON                & Visual Question Answering    & Movies            \\
CEB                  & Character Representation     & Novels          \\
Love and Violence    & Summarization                & TV Shows         \\
\bottomrule
\end{tabular}}
\caption{Overview of 8 more tasks in the character understanding benchmark.}
\label{extended_tasks}
\end{table}

\section{Prompts for GPT4}
\label{prompt}
The following is the prompt we sent to GPT4:
\textbf{For personality traits extraction} 'you are a psychology research assistant designed to help analysis character personalities according to conversations.' \textbf{conversation} 'Here ends the conversation.Give me a character description for each main character according to this conversation.'

\textbf{Character Description}
Here is a character description:Monica: Responsible, caring, and organized. Assertive and confident in her actions.Chandler: Witty and self-deprecating with an approachable sense of humor. Exhibits insecurity and anxiety in his dialogues, making references to uncomfortable situations and questioning his own actions.Rachel: Spontaneous and open to change, she takes risks and is adaptable. She is also reliant on her relationships with others, signifying her dependency and need for support.' 'Joey: Energetic, extroverted, and casual. Lacks the sensitivity of others feelings at times but genuinely care about friends.Phoebe: Quirky, eccentric, and a free spirit. Her train of thought tends to lean toward the unusual and bizarre. However, she is also compassionate and caring.Ross: Insecure and somewhat neurotic and vulnerable. His behavior is indicative of someone who is going through emotional turmoil.'

\section{Name Replacement and Character Comparison}
Here is a single example:\\
Original:\\
 P1: Phoebe, honey, if you hate it so much…
 
Cross-culture replacement:

 P1: Xiaomei, honey, if you hate it so much…
 
Same-culture replacement:

 P1: Susan, honey, if you hate it so much…
Neutral placeholder replacement:

 P1: Speaker A, honey, if you hate it so much…
 
We also provide partial name replacement dictionary in table ~\ref{tab:example}.

\subsection{Cross-Language Uncommon Name Replacement}
With gender swap:[Bojing, Cuixia, Jingjing, Yunsheng, Meilin, Yusong]
Without gender swap:[Cuixia, Bojing,Yunsheng,Jingjing,Yusong, Meilin]

\onecolumn
\begin{table*}[ht]
\centering
\begin{tabular}{|p{4cm}|p{5cm}|p{5cm}|}
\hline
Task name & Before replacement & After replacement \\
\hline
TVShow& Monica, Joey, Chandler, Phoebe, Ross, Rachel &Bojing, Cuixia, Jingjing, Yunsheng, Meilin, Yusong \\
\hline

CSI & Grissom, Catherine, Nick, Sara, Brass & Jingbo, Jingwen, Huaqiang, Lihui, Weimin\\
\hline
PERSONET &Villefort, The Count, Dantès, Caderousse, Danglars, Albert, Mercédès, Elders, Abbé Busoni, Eugénie, Faria priest, Fernand, Franz, Faria, Morre, Monte Cristo & Weina, Nüxia, Danqing, Cailing, Danyu, Ailan, Mengzhe, Lianhua, Busu, Yujun, Fanli, Feiyan, Fangqing, Fangli, Moya, Mingxia\\
\hline
\end{tabular}
\caption{Besides the main characters, there are many names of minor characters and passersby appearing in the tasks. In each task, these secondary and passerby names can be dozens of times more than the number of main characters. Replacing these names is complex but not impossible (e.g., with the help of LLMs). However, to align with the TV show approach of only replacing the names of main characters, we only replaced the main characters in all tasks. From the results, it is clear that even replacing only the main characters has had a significant effect.}
\label{tab:example}
\end{table*}

\twocolumn
replace all the character name \textbf{[Monica, Joey, Chandler, Phoebe, Ross, Rachel]} \\
in the conversation with \\ 
 \textbf{[Bojing, Cuixia, Jingjing, Yunsheng, Meilin, Yusong]}.

\textbf{The comparison before and after name substitution}

The scene overall reflects a group dynamic characteristic of close friends who are comfortable jesting with each other, while also being there in times of crisis or emotional turmoil.\\

1.Monica: Responsible, caring, and organized. Assertive and confident in her actions.\\
1. Bojing: Bojing comes across as practical, level-headed and caring. He often acts as the voice of reason for his friends, attempting to mediate, clarify, and console in various situations. His attempts to play down his date suggest he is a private person who doesn't enjoy sharing intimate details of his life.\\

2. Joey: Energetic, extroverted, and casual. Lacks the sensitivity of others' feelings at times but genuinely care about friends.\\
2. Cuixia: Cuixia seems like a lively, fun-loving character. However, she sometimes shows a more cynical side, quick to suspect something might be wrong with Bojing's date and suggesting a strip joint as a solution to Meilin’s woes.\\

3. Chandler: Witty and self-deprecating with an approachable sense of humor. Exhibits insecurity and anxiety in his dialogues, making references to uncomfortable situations and questioning his own actions.\\
3. Jingjing: Jingjing can be open and candid about his thoughts, even if they seem inappropriate or unusual. He's also humorously self-aware, undercutting his moments of honesty with reminders that he might be oversharing.\\

4. Phoebe: Quirky, eccentric, and a free spirit. Her train of thought tends to lean toward the unusual and bizarre. However, she is also compassionate and caring.\\
4. Yunsheng: Appears offbeat and unusual, suggesting the eating chalk anecdote about his past relationship. He also believes in new-age concepts like auras, showing a more spiritual side.\\

5. Ross: Insecure and somewhat neurotic and vulnerable. His behavior is indicative of someone who is going through emotional turmoil.\\
5. Meilin: Exhibits vulnerability and emotional turmoil, especially regarding his recent divorce. He seems to be fluctuating between hurt, anger, and longing for his past relationship. \\

6. Rachel: Spontaneous and open to change, she takes risks and is adaptable. She is also reliant on her relationships with others, signifying her dependency and need for support.\\
6. Yusong: Yusong presents as impulsive, high-strung and somewhat comical in moments of panic. Fleeing her wedding because of a sudden realization shows she can make drastic decisions based on her emotions.\\

\subsection{Same-Language  Replacement}
With Gender-Matched replacement, we replace all the character name \textbf{[Monica, Joey, Chandler, Phoebe, Ross, Rachel]} \\
in the conversation with \\ 
 \textbf{[Sally, Andrew, Ethan, Olivia, Benjamin, Lauren]}.

With Gender Swap as well, we got  \textbf{[Andrew, Sally,Olivia, Ethan, Lauren,Benjamin]}.

\subsection{Anonymized Speaker Replacement}
replace all the character name \textbf{[Monica, Joey, Chandler, Phoebe, Ross, Rachel]} \\
in the conversation with \\ 
 \textbf{[Speaker 0, Speaker 1, Speaker 2, Speaker 3, Speaker 4, Speaker 5]}.
\label{nameapp}
\textbf{LLM summarized character description}

We use the prompt of "you are a psychology research assistant designed to help analysis character personalities according to conversations........Here ends the conversation.Give me a character description for each main character according to this conversation."
" and we get such result from TVshow Friends scene1.

\section{Benchmark Datasets}
\label{benchdata}
The six selected benchmarks cover a range of character-centric tasks, designed to test different aspects of character understanding(Table \ref{benchmark_datasets}):

\begin{itemize}
    \item \textbf{TVShow Character Guessing}: Based on the TVSHOWGUESS dataset, which contains scripts from five popular TV series: \textit{Friends}, \textit{The Big Bang Theory}, \textit{The Office}, \textit{Frasier}, and \textit{Gilmore Girls}~\cite{sang2022tvshowguess}, this is a multi-choice classification task. The dialogue is anonymized using placeholder identifiers (e.g., P1, P2), and the model must predict the correct speaker for each line, testing its ability to identify characters based on context.

    \item \textbf{Screenplay}: Using screenplays from nine popular movies, such as \textit{Avengers: Endgame} (2019), \textit{Dead Poets Society} (1989), \textit{John Wick} (2014), and \textit{The Shawshank Redemption} (1994)~\cite{baruah2023character}, this is a coreference resolution task. The model is tasked with linking character mentions throughout the text, forming coreference clusters and resolving ambiguities in character references.

    \item \textbf{Personality Understanding}: Using the PERSONET dataset, which comprises approximately 32,000 samples from 33 classic literature books~\cite{yu2023personality}, this task involves predicting the most likely personality trait depicted in a given text snippet, along with its preceding context. Formulated as a multi-choice problem with distractor traits, this task evaluates the model’s ability to infer character personalities from descriptive text.

    \item \textbf{CSI Role Detection}: Based on 39 episodes of the TV series \textit{CSI}~\cite{frermann_whodunnit_2018}, this sequence labeling task requires the model to identify mentions of the perpetrator in dialogues and scene descriptions, focusing on role detection within crime-related narratives.

    \item \textbf{FriendsQA}: Derived from 1,222 scenes from the first four seasons of \textit{Friends}~\cite{yang_friendsqa_2019}, this dataset supports an open-domain question answering task. The model is given a dialogue excerpt and a manually crafted question and must extract the specific span in the dialogue that answers the question, testing its comprehension of dialogue context.

    \item \textbf{SUMMSCREEN}: Built from transcript-recap pairs of 91 TV series collected from community sources such as The TV MegaSite, Inc. (TMS) and ForeverDreaming (FD)~\cite{chen_summscreen_2022}, this dataset supports a summarization task. The model is required to generate a coherent summary from dialogue utterances and scene descriptions, capturing key plot points while excluding irrelevant details.
\end{itemize}
\section{Gist Prompting Strategies Across Tasks}
\label{gistprompts}
We design \textbf{Gist Prompting} methods to reduce reliance on memorization and promote inference-based character understanding. Below, we detail the task-specific prompt formulations.

\subsection{Task-Specific Gist Prompts}
\paragraph{Character Guessing} 
\textit{"Analyze the dialogue and infer the most likely speaker based on their speech patterns, personality traits, and relationships with other characters. Do not rely on direct recall of specific names but rather use reasoning from the provided context."}

\paragraph{Coreference Resolution} 
\textit{"Identify which mentions in the text refer to the same character by analyzing linguistic patterns, contextual cues, and discourse coherence. Do not rely on pre-learned entity associations but rather resolve coreference based on narrative structure."}

\paragraph{Personality Understanding} 
\textit{"Based on the character’s dialogue and actions, infer their personality traits. Focus on how they interact with others, their tone, and behavioral tendencies. Avoid using prior memorized descriptions of characters and derive insights purely from the given passage."}

\paragraph{Role Detection} 
\textit{"Determine whether a character plays a key role in the narrative (e.g., antagonist, victim, or protagonist) by analyzing their function within the story. Identify how their actions and relationships define their role rather than relying on predefined labels."}

\paragraph{Open-Domain QA} 
\textit{"Answer the question using only the given passage. Identify relevant information by reasoning over the text and analyzing interactions between characters. Do not rely on external knowledge or prior memorized facts."}

\paragraph{Summarization} 
\textit{"Summarize the passage by identifying key events, character interactions, and narrative progression. Construct a coherent summary that captures the essence of the scene without directly copying text segments."}

\section{Human Study}
\label{human}

While our primary focus is to isolate memorization from reasoning in LLMs, 
we ran a small-scale human study to verify that our name-replacement strategy 
does not remove essential reasoning clues and that gold answers remain valid. 
We implemented two complementary experiments: Exp.~1 evaluates objective accuracy; 
Exp.~2 evaluates subjective semantic integrity and gold-label validity.

\paragraph{Participants.}
Three adult participants were recruited: two had never watched the target TV series, 
and one had prior familiarity. All responses were anonymous; no risks were involved.

\paragraph{Materials.}
A set of 100 multiple-choice items (randomly sampled from our benchmark) was used. 
Each item contains a short excerpt and four options with a pre-defined gold label.

\paragraph{Exp.~1: Procedure (Objective Accuracy).}
Each participant solved the same 100 items twice: first in the \emph{Original} condition 
(with canonical character names), and then in the \emph{Name-Replaced} condition 
(with consistent substitutions). After the second round, participants re-examined 
the items and flagged any case where they believed name replacement removed necessary 
reasoning information and thus rendered the item unsolvable.

\paragraph{Exp.~1: Instructions.}
“You will be presented with multiple-choice questions based on short excerpts from a TV script. 
Please answer each question to the best of your ability. After completing both rounds, 
review the questions again and mark any cases where the replacement of names 
made the question unsolvable.”

\paragraph{Exp.~1: Results.}
Table~\ref{tab:humanstudy} reports accuracy before and after name replacement and the number of flagged items.
Performance differences were small for all participants (Person~1: $\Delta=+0.9$ pp; 
Person~2: $\Delta=-2.7$ pp; Person~3: $\Delta=-7.2$ pp; max $|\Delta|=7.2$ pp). 
No item was flagged as unsolvable.

\begin{table}[h]
\centering
\begin{tabular}{lccc}
\toprule
\textbf{Participant} & \textbf{Accuracy (Original)} & \textbf{Accuracy (Replaced)} & \textbf{Flagged Items} \\
\midrule
Person 1 (unfamiliar) & 46.2\% & 47.1\% & 0 \\
Person 2 (unfamiliar) & 60.7\% & 58.0\% & 0 \\
Person 3 (familiar)   & 86.6\% & 84.4\% & 0 \\
\bottomrule
\end{tabular}
\caption{Human study (Exp.~1) results before and after name replacement.}
\label{tab:humanstudy}
\end{table}

\paragraph{Exp.~2: Procedure (Semantic/Gold Integrity).}
To directly test whether name replacement affects gold-answer validity, 
participants were shown, for each item, the \emph{Name-Replaced} text alongside the \emph{gold option}. 
They then judged whether the replacement (i) \emph{changed} the correct answer or 
(ii) made the answer \emph{ambiguous} given the excerpt.

\paragraph{Exp.~2: Instructions.}
“For each item, you will see the excerpt with replaced names and the official correct option (gold). 
Please indicate whether the name replacement changes the correct answer or makes it ambiguous. 
If neither applies, select ‘no issue.’”

\paragraph{Exp.~2: Results.}
Across $3 \times 100 = 300$ item–judgments, \emph{no} case was flagged as “changed” or “ambiguous.” 
This indicates that name replacement neither alters gold labels nor compromises their interpretability.

\paragraph{Discussion.}
Exp.~1 shows that human accuracy is stable under name replacement with no unsolvable items, 
indicating preserved reasoning clues. Exp.~2 shows that gold answers remain valid and unambiguous 
after replacement. Together, these results support that our intervention suppresses 
verbatim memorization while preserving the semantic structure required for inference.

\section{Prompt Robustness Under Minimal Paraphrases}
\label{app:prompt-robust}

\paragraph{Goal.}
To test whether our main conclusions are sensitive to prompt wording, we evaluated
\emph{minimal paraphrases} of the two prompting conditions (Verbatim vs.\ Gist).
The aim is to keep task framing and structure unchanged while slightly altering phrasing,
and to check whether the accuracy pattern in the main paper persists.

\paragraph{Design.}
For each condition we authored two paraphrastic variants (V1/V2 for Verbatim, G1/G2 for Gist),
keeping the shared base instruction:
\textit{``The following is a dialogue. Please identify who they are. Choose the correct option.''}
For every model and variant we ran three trials with the same decoding settings as in the main
experiments (temperature, top-$p$, max tokens), and report mean accuracy over the three runs.

\paragraph{Prompt texts (minimal variants).}
\textbf{Verbatim–V1:} \textit{``Answer by \underline{direct memorization retrieval}. Do not infer
from traits, interactions, or context. Recall exact names from prior dialogue knowledge.
Output the final option only; no explanation.''}
\quad
\textbf{Verbatim–V2:} \textit{``Use \underline{literal recall} of prior dialogue. Do not rely on
reasoning about relationships or events. Output only the letter of the matching name; no steps.''}

\noindent
\textbf{Gist–G1:} \textit{``Decide by analyzing \underline{relationships, key events, and traits};
match these cues to known characters. Do not directly retrieve who spoke from memory.
Output the final option only.''}
\quad
\textbf{Gist–G2:} \textit{``Rely on \underline{semantic cues} (who relates to whom, what happened,
stable traits). Avoid direct memorization of names or lines. Provide only the chosen option.''}

\paragraph{Results.}
Across all models, the qualitative ordering from the main paper is preserved
(\texttt{Baseline} $\ge$ \texttt{Verbatim} $\ge$ \texttt{Gist} for GPT\!-\!4o and LLaMA3.3\!-\!70B;
for Deepseek V3, \texttt{Verbatim} remains slightly above \texttt{Baseline}, with \texttt{Gist}
lowest). Within-condition differences between paraphrases are small (typically within
$\approx$0.3–0.7 pp), indicating low prompt sensitivity under minimal wording changes.

\begin{table}[h]
\centering
\begin{tabular}{lcccccc}
\toprule
\multirow{2}{*}{\textbf{Model}} &
\multicolumn{1}{c}{\textbf{Baseline}} &
\multicolumn{2}{c}{\textbf{Verbatim}} &
\multicolumn{2}{c}{\textbf{Gist}} \\
\cmidrule(lr){3-4} \cmidrule(lr){5-6}
& mean & V1 mean & V2 mean & G1 mean & G2 mean \\
\midrule
GPT\!-\!4o        & 78.23 & 77.32 & 77.74 & 70.48 & 69.82 \\
LLaMA3.3\!-\!70B  & 60.19 & 57.10 & 57.62 & 52.22 & 51.63 \\
Deepseek\ V3      & 56.41 & 58.30 & 58.86 & 49.32 & 49.87 \\
\bottomrule
\end{tabular}
\caption{Minimal paraphrase robustness. Numbers are mean accuracy (\%) over three runs per variant.
The qualitative ordering matches the main results; within-condition differences are small.}
\label{tab:minimal_paraphrase}
\end{table}

\end{document}